\documentclass{article}

\usepackage{arxiv}

\usepackage[utf8]{inputenc} 
\usepackage[T1]{fontenc}    
\usepackage{hyperref}       
\usepackage{url}            
\usepackage{booktabs}       
\usepackage{amsfonts}       
\usepackage{nicefrac}       
\usepackage{microtype}      
\usepackage{graphicx}
\usepackage{cite}

\title{
    LipidBERT: A Lipid Language Model Pre-trained on \\ METiS \textit{de novo} Lipid Library
    }

\author{
    \textbf{Tianhao Yu}\thanks{\textbf{Corresponding authors. For collaborations or trials using our \textit{de novo} lipid library, LipidBERT model, or PhatGPT model, please contact Tianhao Yu and Kai Wang via email at thyu@metispharma.com and kwang@metispharma.com, respectively.} \\ \\
    \textbf{Author Contributions:} \\
    Tianhao proposed, designed, and implemented the AI models, and wrote the manuscript. Kai proposed, designed, and supervised the AI models and the virtual lipid library, and also wrote and reviewed the manuscript. Cai designed and developed the virtual lipid library. Zhourui, Feng, Lin, and Andong provided the wet-lab experimental data for fine-tuning. Kangjie and Xuan assisted with the development of MD/QM descriptors. Xicheng developed the front-end user interface of our AiLNP platform. Jiali contributed to the design of the AI models. Wenshou and Chris reviewed the manuscript and organized the dry-wet lab collaboration.}
    \hspace{2em}
    \textbf{Cai Yao}\hspace{2em}
    \textbf{Zhuorui Sun}\hspace{2em}
    \textbf{Feng Shi}\hspace{2em}
    \textbf{Lin Zhang}\hspace{2em}
    \textbf{Kangjie Lyu}\\\\
    \textbf{Xuan Bai}\hspace{2em}
    \textbf{Andong Liu}\hspace{2em}
    \textbf{Xicheng Zhang}\hspace{2em}
    \textbf{Jiali Zou}\hspace{2em}
    \textbf{Wenshou Wang}\hspace{2em}
    \textbf{Chris Lai}\\\\
    \textbf{Kai Wang}\footnotemark[1]\\
    \\\\
    METiS Pharmaceuticals\\
    \\\\
    \texttt{\{thyu, cyao, zrsun, feng.shi, lzhang, kjlv, xbai, adliu, xczhang, jlzou,}\\
    \texttt{wwang, CLai, kwang\}@metispharma.com}
}

\begin{document}
\maketitle

\begin{abstract}
Ionizable lipids play a critical role in defining and modulating the properties of lipid nanoparticles (LNPs). Transformer-based architectures have demonstrated significant potential in natural language processing (NLP) and self-supervised learning, particularly by reducing the reliance on extensive labeled data. However, unlike small molecules, which benefit from numerous public databases, the publicly available ionizable lipid structures remain insufficient for effective NLP pre-training. In this study, we address this gap by generating and maintaining a database of 10 million virtual lipids through METiS's in-house \textit{de novo} lipid generation algorithms and lipid virtual screening techniques. These virtual lipids serve as a corpus for pre-training, lipid representation learning, and downstream task knowledge transfer, culminating in state-of-the-art LNP property prediction performance. We propose LipidBERT, a BERT-like model pre-trained with the Masked Language Model (MLM) and various secondary tasks. Additionally, we compare the performance of embeddings generated by LipidBERT and PhatGPT, our GPT-like lipid generation model, on downstream tasks. The proposed bilingual LipidBERT model operates in two languages: the language of ionizable lipid pre-training, using in-house dry-lab lipid structures, and the language of LNP fine-tuning, utilizing in-house LNP wet-lab data. This dual capability positions LipidBERT as a key AI-based filter for future screening tasks, including new versions of METiS \textit{de novo} lipid libraries and, more importantly, candidates for \textit{in vivo} testing for orgran-targeting LNPs. To the best of our knowledge, this is the first successful demonstration of the capability of a pre-trained language model on virtual lipids and its effectiveness in downstream tasks using web-lab data. This work showcases the clever utilization of METiS's in-house \textit{de novo} lipid library as well as the power of dry-wet lab integration.
\end{abstract}

\newpage
\section{Introduction}
mRNA vaccines, comprising of a messenger RNA encapsulated in lipid nanoparticles (LNPs) have shown promising results in preventing COVID-19 infection, as demonstrated by Pfizer-BioNTech (BNT162b2) and Moderna (mRNA-1273)~\cite{sahin2020covid,jackson2020mrna,suzuki2021difference,schoenmaker2021mrna,sun2023structure,fang2022advances}. The fragile nature and degradability issue of mRNA molecules make it challenging to deliver them to target cells and tissues effectively. LNPs have gained significant attention as an effective and adaptable delivery system for mRNA therapeutics. LNPs have demonstrated their ability to encapsulate RNA molecules effectively, shielding them from degradation and enabling cellular entry to perform their therapeutic roles. Additionally, LNPs can bypass several biological barriers, promoting endosomal escape, cellular uptake, and drug release, thus enhancing the threapeutic efficacy of mRNA. Consequently, LNPs have become a popular platform for mRNA therapeutics development, especially in vaccines and gene therapies. Typical LNP formulations include ionizable lipids, phospholipids, cholesterol, and PEG-lipids. Ionizable lipids play a crucial role in the formulation and functionality of LNPs, which are pivotal in the delivery of mRNA delivery. The fundamental distinction between cationic and ionizable lipids resides in the presence of a secondary or tertiary amine group within the hydrophilic head, which allows for charge modulation in response to varying pH levels.

With the rapid development of machine learning techniques and computational power, deep learning has become increasingly applied in molecular property prediction. For small molecules, 
graph neural networks (GNNs) have exhibited significant potential in a number of benchmark tasks ~\cite{duvenaud2015convolutional,kearnes2016molecular,kipf2016semi,gilmer2017neural,coley2017convolutional,yang2019analyzing}. Recently, the Transformer architecture~\cite{vaswani2017attention} has been widely recognized as the most powerful neural network in the field of NLP. Although the Transformer was originally designed for processing sequential data, there have been several attempts to apply this architecture in GNNs~\cite{rong2020GROVER,ying2021graphomer,rampavsek2022graphgps}.

Apart from graph representation of molecules, molecular SMILES is another representation form of sequential data, making it naturally suitable for training on Transformer-like models. ChemBERTa~\cite{chithrananda2020chemberta,ahmad2022chemberta2} leverages the large amount of available SMILES data (up to 77M) for small molecules as a corpus for pre-trainining a RoBERTa~\cite{liu2019roberta} model. However, unlike small molecules, pre-training on ionizable lipids SMILES is significantly hindered by the scarcity of experimentally confirmed structures. The AGILE platform~\cite{xu2023agile} provides approximately 60,000 ionizable lipid structures, yet this number is not comparable to the magnitude of available small molecule structures. Furthermore, the conventional combinatorial chemistry nature of the AGILE platform makes the resulting lipid structures less diverse and hence not a suitable corpus for lipid structure representation learning.

In this paper, we present LipidBERT, a model pre-trained on an \textit{in silico} generated and meticulously screened lipid library constructed via fragment-based generative methods and reinforcement learning (RL). Molecular fragments with both lipid and small molecule features were generated by RL and connected to form ionizable lipids. These were then subjected to a series of knowledge-, molecular dynamics-, and AI-based filters. The resulting library consists of 10 million ionizable lipids, which were subsequently used for pre-training several BERT-like~\cite{devlin2018bert} models on different data scales. The models were then fine-tuned for downstream LNP property prediction tasks, using wet-lab labeled data from our experimental team. The fine-tuned prediction model exhibits a remarkably high Pearson correlation coefficient (~0.8), demonstrating the power of pre-trained lipid language models on virtual datasets. Apart from language models, we also have graph-based and several distribution-based models, which might be reported in the near future.

\section{Methods \& Results}

\subsection{METiS \textit{de novo} Lipid Library}

The details of our \textit{de novo} lipid generation algorithms are considered trade secrets and are beyond the scope of this study. However, we will briefly describe METiS's AI technologies, as they form the foundation of the lipid library and lipid language model. The goal is to develop an algorithm that can generate ionizable lipids that are both potent and structurally novel and diverse. The first set of METiS \textit{de novo} ionizable lipids was generated using a fragment-based method, leveraging the structures of publicly available ionizable lipids from literature and patents, and small molecules from open-source libraries, particularly clinically verified drugs. A fully automated fragmentation algorithm cuts an ionizable lipid into 7-12 pieces, some of which are optional. Each fragment is clearly defined in terms of length, atom types, connection patterns, and other characteristics. Reinforcement learning models, trained on fragment libraries constructed from existing ionizable lipids and small molecules, combined with a sequence-based generative method, are implemented to generate all building blocks with fragment-specific knowledge-based reward functions. These fragments are then reconnected using an in-house connecting algorithm. This approach ensures that the generated fragments are not only diverse but also optimized for their specific roles in the final ionizable lipid structure. 

Following the generation process, the generated molecules undergo a stringent screening process. Knowledge-, AI-, and molecular dynamics-based filters are implemented to screen the large pool of generated lipids. By rigorously applying the knowledge-based filters, the process effectively filters out unreasonable or impractical molecular structures. LipidBERT-based prediction models serve as AI-based filters, while multiple in-house molecular dynamics-based filters ensure that the lipids behave well in simulated lipid bilayers, self-assembled lipid nanoparticles, LNP-endosome system to understand endosomal escape and other relevant biological systems. The details of the METiS in-house molecular dynamics platforms are not discussed here since they are also beyond the scope of the current study. We will publish studies showcasing our molecular dynamics capabilities including high-throughput all-atom LNP bilayer simulations, LNP self-assembly, LNP endosomal escape, and protein corona simulations in the future.

Other distribution-based and sequence-based generative methods are then used to generate more lipids, further enlarging and diversifying the lipid library based on the first generation of the METiS lipid library. The lipid library and the LipidBERT-based prediction models are updated monthly with new model updates and newly generated wet-lab data. 

\subsection{Lipid Generation Models, METiS \textit{de novo} Lipid Library and Lipid/LNP Prediction Models}

\begin{figure}[ht]
    \centering
    \includegraphics[width=13cm]{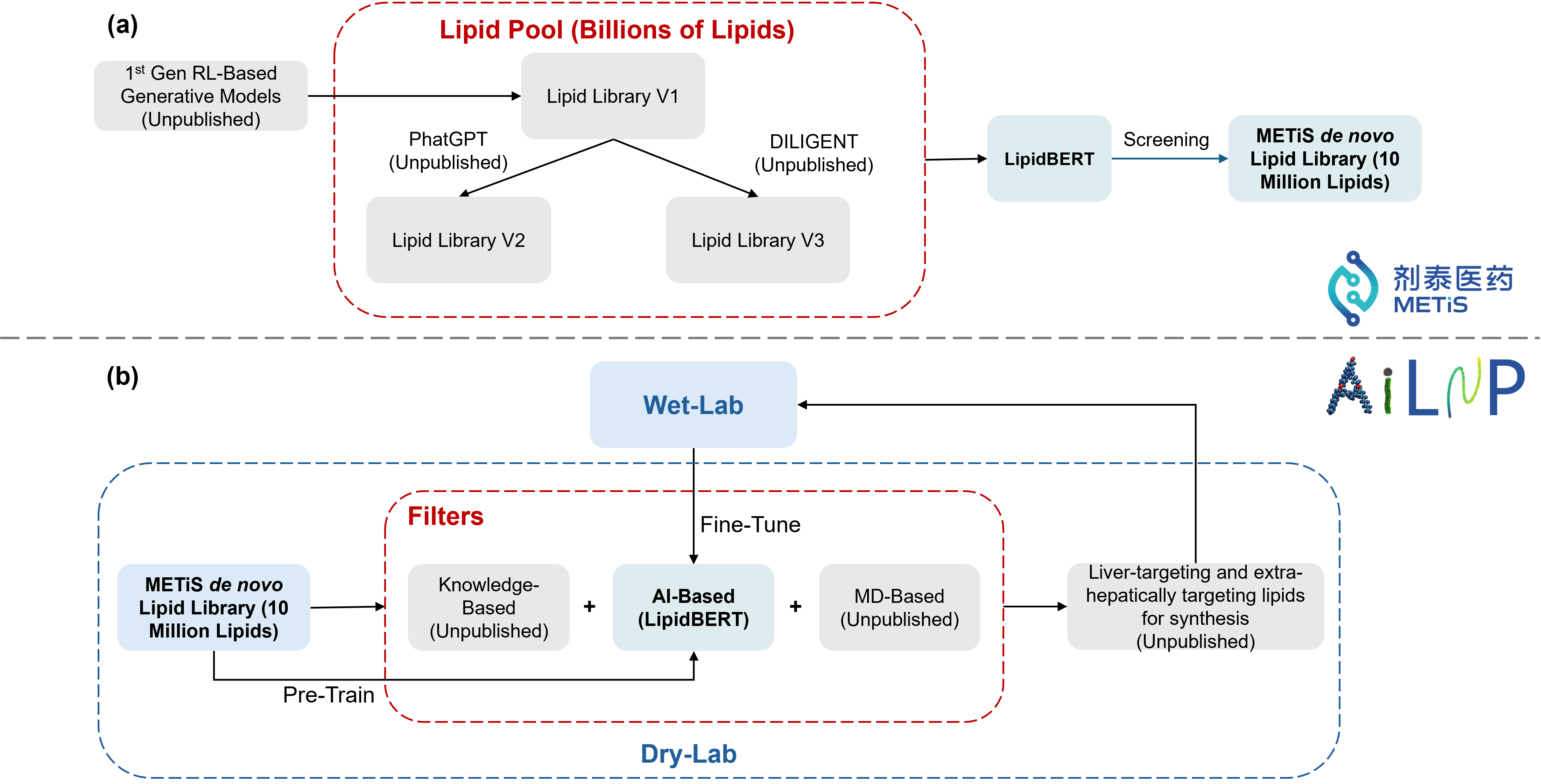}
    \caption{
        Schematic representation of (a) the METiS \textit{de novo} lipid library facilitator, and (b) the real-world lipid filter. "Unpublished" refers to models and experimental methods/results that may be published in the future and has not been discussed in details in this study.
        }\label{fig:library_in_vivo}
\end{figure}

LipidBERT plays a pivotal role in METiS's AI-driven LNP delivery strategy, serving two key functions: facilitating the METiS \textit{de novo} lipid library and acting as a real-world lipid filter. These roles are illustrated in Figure \ref{fig:library_in_vivo}.

\subsubsection{METiS \textit{de novo} Lipid Library Facilitator}

As previously mentioned, RL-based lipid generation algorithms produced the first version of the METiS \textit{de novo} Lipid Library. These algorithms incorporated extensive human knowledge and understanding, as we began constructing the library from scratch with only tens of thousands of existing ionizable lipid structures available from literature and patents—far too few to support any meaningful language-based models. Transformer-based sequence understanding techniques enabled us to decode lipid sequences, while reinforcement learning-based generative methods empowered us to precisely design reward functions that effectively conveyed our expert insights into lipids and LNPs. Subsequent versions of the lipid generation methods have built upon this initial version. They utilized the distribution or corpus of the first-version lipid library and employed distribution-based and GPT-like generative methods. These advanced lipid generation models now yield billions of lipids each month. LipidBERT, a model fluent in the language of lipids through pre-training on lipid structures and the language of LNPs through fine-tuning with real LNP data, is the optimal facilitator for dynamically updating the METiS \textit{de novo} Lipid Library. We maintain this library at a constant size of 10 million lipids, renewing it at least monthly based on updates to the lipid generation algorithms and the addition of new wet-lab data.

\subsubsection{Real-World Lipid Filter}

Virtual lipids remain hypothetical until they are positively predicted by \textit{in silico} methods and subsequently validated through \textit{in vivo} testing. However, the number of lipids we can test each month is limited. LipidBERT serves as a critical real-world lipid filter, alongside other unpublished filters such as MD-based filters and knowledge-based filters, to determine which lipids can transcend their virtual existence and be evaluated in the real world.

\subsection{Pre-Training}
In the vanilla BERT~\cite{devlin2018bert} model, two tasks are performed simultaneously: the Masked Language Model (MLM) and Next Sentence Prediction (NSP). In MLM, a specified number of tokens are randomly selected from the input sequence to be masked, and the model attempts to predict these masked tokens based on the other tokens. For NSP, two sentences are concatenated and separated by a separation token, and the model predicts whether the latter sentence is the actual next sentence of the previous one.

To integrate the capabilities of the vanilla BERT model into our LipidBERT model, we retain the MLM task in every model but design different secondary tasks during pre-training, as the NSP task does not naturally fit into BERT models trained on molecular SMILES. The MLM is combined with different secondary tasks: number of tails prediction, connecting atom prediction - sequence/token classification, head/tail classification, rearranged/decoy SMILES classification, and all possible combinations of the above secondary tasks. Head/tail and connecting atom are the main structural features that distinguish lipid from small molecules. These secondary tasks force the model to gain deeper understanding of the lipid structure. All models were trained on different data scales, including randomly sampled sets of 0.05 million, 0.25 million, 1 million, 3 million, 5 million, and all 10 million ionizable lipid SMILES from our library. A character tokenizer was used, and each sequence was padded to the maximum length of 128, except for models involving rearranged/decoy smiles classification, which requires two molecular SMILES and thus has a maximum length of 256. Note that we have a larger model that can process single SMILES sequences up to a length of 256, but here we focus on our base model. All SMILES were converted into canonical form using RDKit~\cite{landrum2006rdkit}. 

The MLM settings remain the same for all models, using the vanilla BERTbase configuration. The model has 12 encoder blocks and in each block, each self-attention layer returns a vector of size 768, which is then fed into a fully connected network. The hidden layers of the fully connected network output a vector of size 3072 and the final layer returns a vector of size 768. A 12-head attention mechanism is applied in the self-attention calculation. We use a batch size of 128 and all models are trained for 10 epochs using the AdamW~\cite{loshchilov2017adamw} optimizer with an adaptive learning rate starting from 5e-5.

\begin{figure}[ht]
    \centering
    \includegraphics[width=13cm]{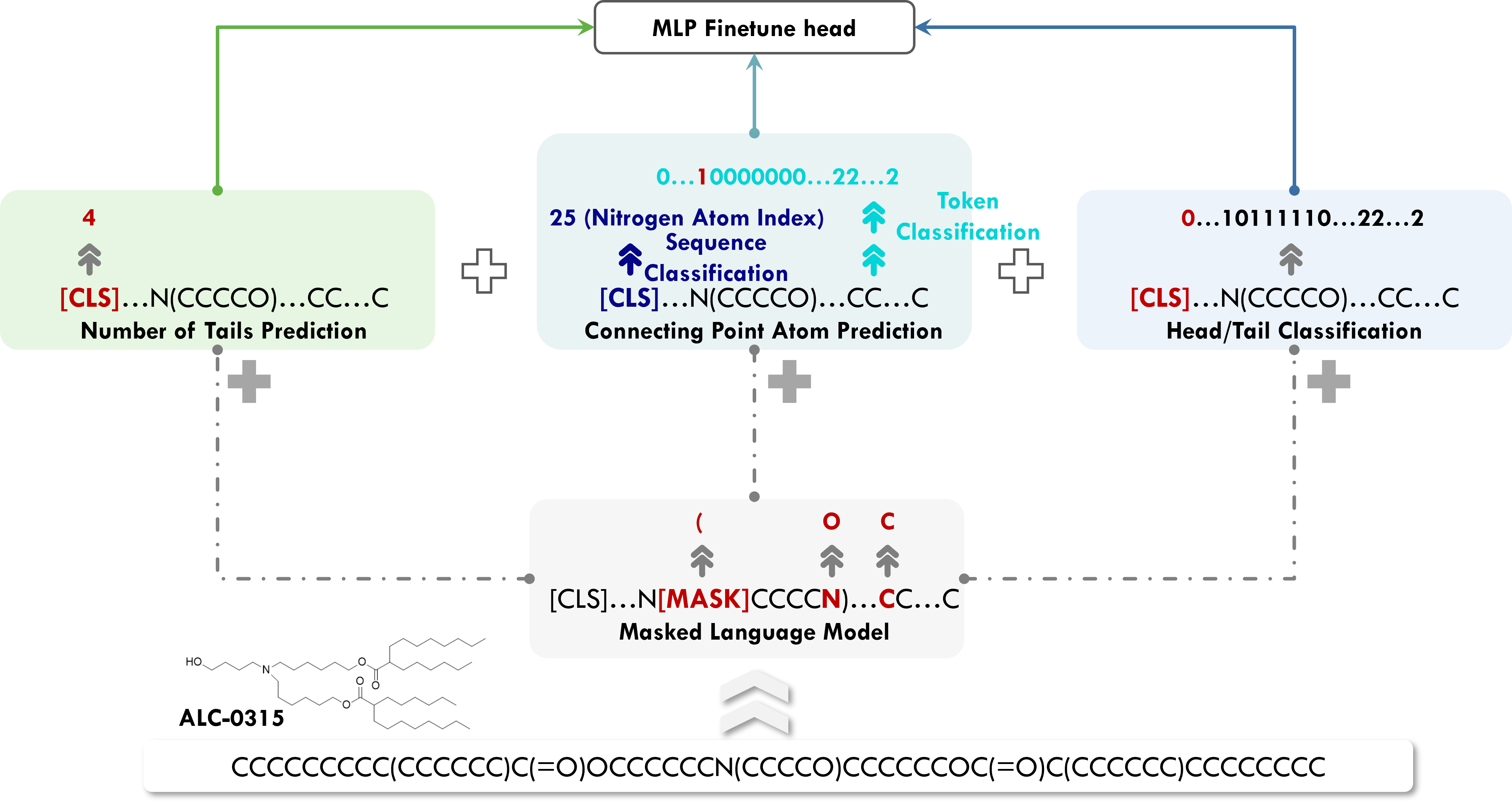}
    \caption{
        Schematic representation of the Masked Language Model (MLM) and various secondary tasks, including Number of Tails Prediction, Connecting Atom Prediction - Sequence/Token Classification, and Head/Tail Classification.
        }\label{fig:BERT}
\end{figure}

\subsubsection{Masked Language Model (MLM)}
The processing of sequences in the Masked Language Model (MLM) remains largely consistent with the MLM model in the vanilla BERT model, as illustrated in Figure~\ref{fig:BERT}. 15\% of tokens are randomly selected for each SMILES sequence. If one token is selected, it is replaced in the following ways: 80\% of the time with the [MASK] token, 10\% of the time with a random token, and 10\% of the time it remains unchanged. It should be noted that the selected 15\% tokens are not limited to tokens representing atoms (e.g., N, C, O), and may include non-atomic tokens such as brackets. All selected tokens will contribute to the cross-entropy loss via predicting their original tokens.

\begin{figure}[ht]
    \centering
    \includegraphics[width=12cm]{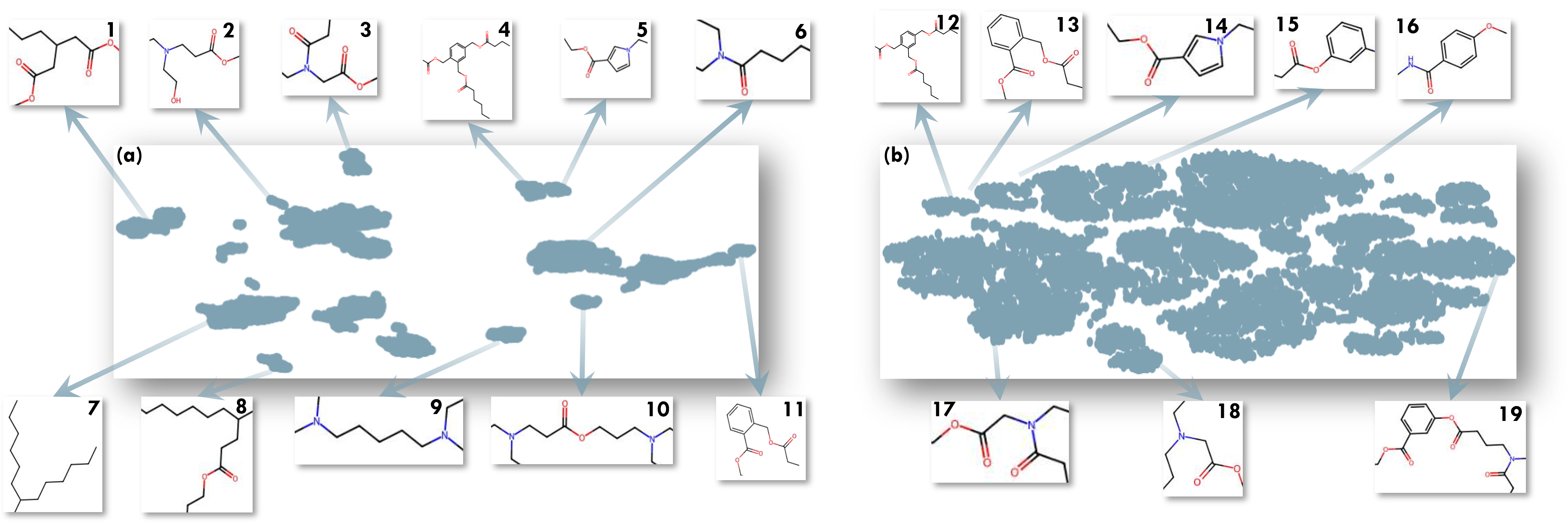}
    \caption{
        Projected embeddings on 2D from the 768-dimensional [CLS] embeddings generated via the pre-trained Masked Language Model. Dimensionality reduction was performed using (a) UMAP~\cite{mcinnes2018umap}, and (b) t-SNE~\cite{van2008tsne}.
        }\label{fig:MLM_embedding}
\end{figure}

Figure~\ref{fig:MLM_embedding} illustrates the visualization of 20,000 randomly selected lipid embeddings from the training set. The [CLS] embedding is reduced from 768 to 2 dimensions using (a) UMAP~\cite{mcinnes2018umap} and (b) t-SNE~\cite{van2008tsne}. It is evident that the two methods generate quite different dimensionality reduction results. UMAP yields much more closely clustered groups of points than t-SNE. The clustered points exhibit common or similar molecular substructures, which are explicitly indicated in the figure. For instance, in Figure~\ref{fig:MLM_embedding} (a), all lipids in group 8 contain at least one branched alkyl chain. Some groups with similar shared substructures are positioned close to each other, such as groups 9 and 10. Group 9 features an alkyl chain connecting two nitrogen atoms, while group 10 has an additional ester group on the alkyl chain. Another example is groups 4 and 5, where group 4 has a benzene ring connecting three ester groups, and group 5 has a five-membered ring connecting one ester group. Similarly, in Figure~\ref{fig:MLM_embedding} (b), groups 12, 13, and 15 are projected close to each other, all containing a benzene ring connected to 3, 2, 1 ester groups, respectively. Group 14 lies close to groups 12, 13, and 15 and has a five-membered ring connected to an ester group. Both figures demonstrate that the model can categorize lipids solely based on their SMILES, even in a condensed 2D space.

\subsubsection{Number of Tails Prediction}
In our lipid dataset, the number of tails of all lipids falls in the range from 2 to 6, resulting in a 5-class sequence classification problem. A linear classifier is added on top of the model and the cross entropy loss is combined with the MLM loss to contribute to the total loss. In the given example of ALC-0315 lipid in Figure~\ref{fig:BERT}, four tails can be identified, thus the classification label is 4.

\begin{figure}[ht]
    \centering
    \includegraphics[width=10cm]{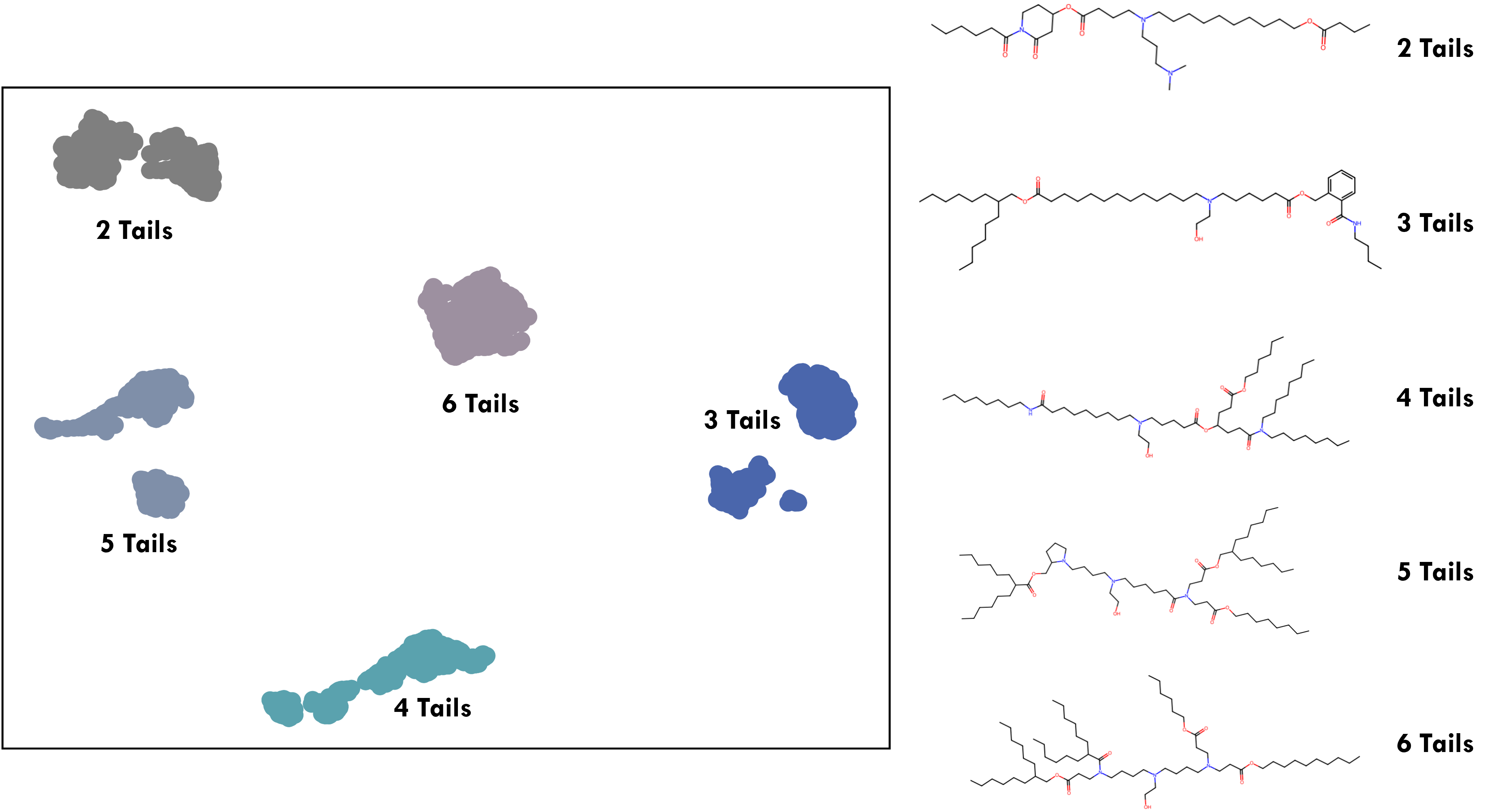}
    \caption{
        Projected embeddings on 2D from the 768-dimensional [CLS] embeddings generated via the Masked Language Model + Number of Tails model. Dimensionality reduction was performed using UMAP~\cite{mcinnes2018umap}.
        }\label{fig:Ntails}
\end{figure}

The projected embeddings via dimensionality reduction using UMAP~\cite{mcinnes2018umap} is presented in Figure~\ref{fig:Ntails}. A total of 800 lipids were randomly selected from the training set for each group based on the number of tails, specifically from 2 to 6 tails, resulting in 4,000 sampled lipids. The number of tails labels were explicitly calculated for the lipids in the training set by counting the non-branching chains at the tips of the tail chains. Examples of lipids with 2 to 6 tails are also provided in the figure, selected for their high structural diversity but poor AI-predicted LNP properties. The figure indicates that the pre-trained model can perfectly distinguish lipids based on the number of tails, as evidenced by the clear separation and clustering of lipids with the same number of tails. These results suggest that the secondary task of predicting the number of tails during the pre-training process successfully learned this feature from lipid smiles. It should also be noted that separate closely positioned subclusters can be observed within the 2-, 3-, 4-, and 5-tail groups. In the 2, 4, and 5-tail groups, the subclusters are separated by the model due to structural deviations, where one subcluster contains ring-like structures and the other does not. In the 3-tail groups, the ring-containing groups are further divided into two subgroups, where the lipids in the smaller group share nitrogen-containing rings and the larger group contains regular nitrogen-free rings. This indicates that the model retains some structural differentiation capabilities due to the Masked Language Model task in its pre-training process.

\subsubsection{Connecting Atom Prediction - Sequence Classification}
We define the connecting atom as the single atom that links the head and tail fragments of a lipid. Two methods are proposed for prediction - sequence and token classification. In sequence classification, predicting the connecting atom involves identifying its atomic index. Summarizing the connecting atomic indices in the our dataset results in 51 individual indices, thus creating a 51-class classification problem. Similar to the Number of Tails Prediction model,
a linear classifier is appended on top of the model for sequence classification. Figure~\ref{fig:BERT} presents an example of predicting the connecting point for ALC-0315: the nitrogen atom (with an atomic index of 25 in its SMILES) links its head and tail fragments. Therefore, the prediction label for ALC-0315 SMILES is 25.

\begin{figure}[ht]
    \centering
    \includegraphics[width=10cm]{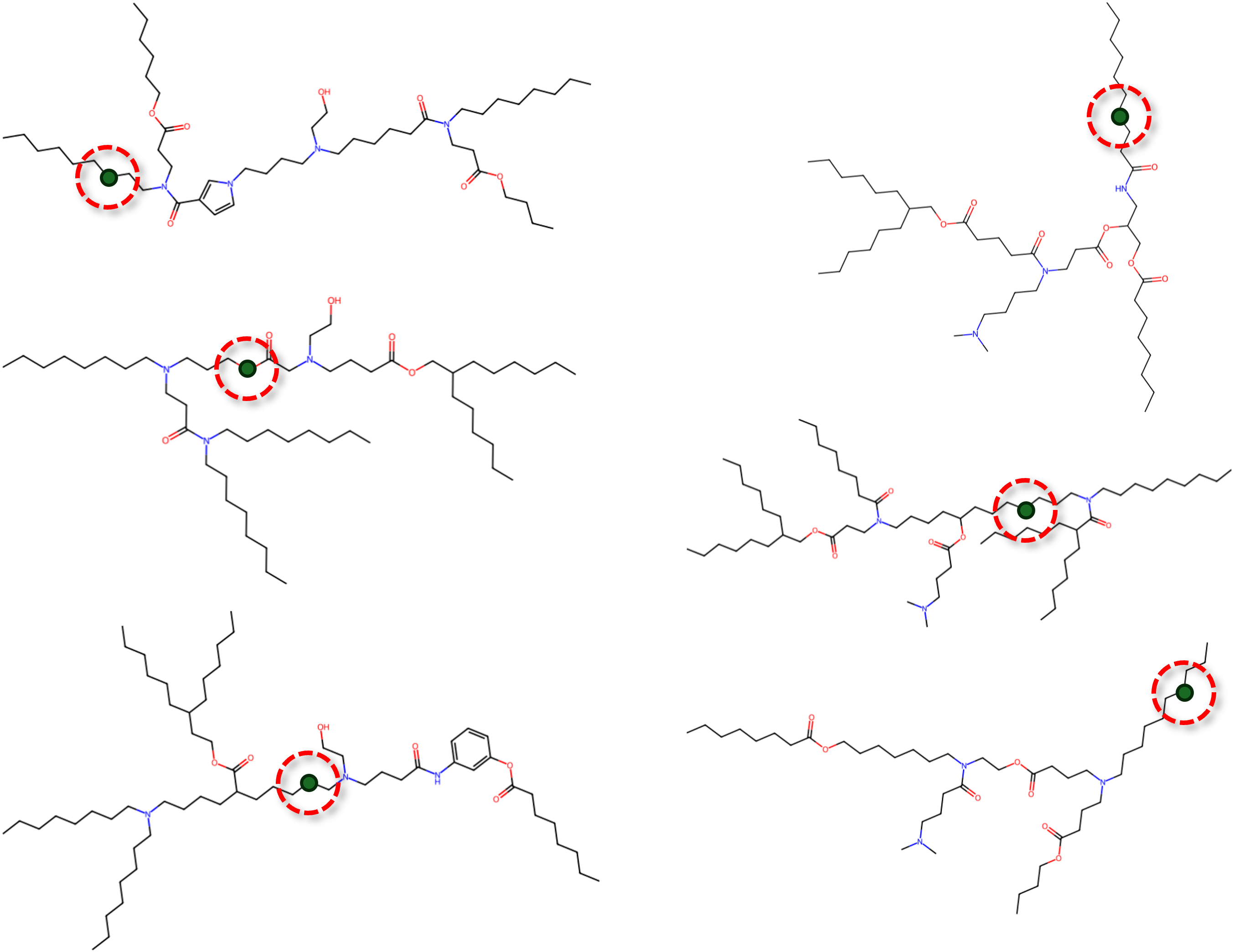}
    \caption{
        Visualization of the predicted connecting atom between head and tail via sequence classification, using the connecting points prediction head in the pre-trained model. The predicted atoms are explicitly marked. The visualization was created using RDKit~\cite{landrum2006rdkit}. We intentionally selected six lipids with low AI-predicted values but high diversity from the sampled set.
        }\label{fig:connecting_points}
\end{figure}

Predicting the connecting atom via sequence classification is one of the most challenging training objectives. This task involves a 51-class classification problem, 
as our database includes 51 individual connecting atom indices. Classifying the correct connecting atom from 51 possible classes for each lipid is significantly more difficult than other secondary tasks. The results are depicted in Figure~\ref{fig:connecting_points}. Six lipids were randomly sampled from the training set, and the predicted connecting points are explicitly marked. It can be observed that the pre-trained model fails to identify the correct connecting point for all sampled lipids. These results highlight the difficulty of the connecting point prediction secondary task.

\subsubsection{Connecting Atom Prediction - Token Classification}
Unlike the previously mentioned secondary tasks that involve sequence classification, the connecting atom token classification focuses on classifying each token individually to determine if it represents a connecting atom. Consequently, the labels for a molecule form a vector of zeros, with only one token assigned a classification label of 1, as illustrated in Figure~\ref{fig:BERT}). To facilitate this classification, a classification head is utilized to transform the 768-dimensional embedding of each token into a vector of length 2 for 0/1 classification.

\begin{figure}[ht]
    \centering
    \includegraphics[width=10cm]{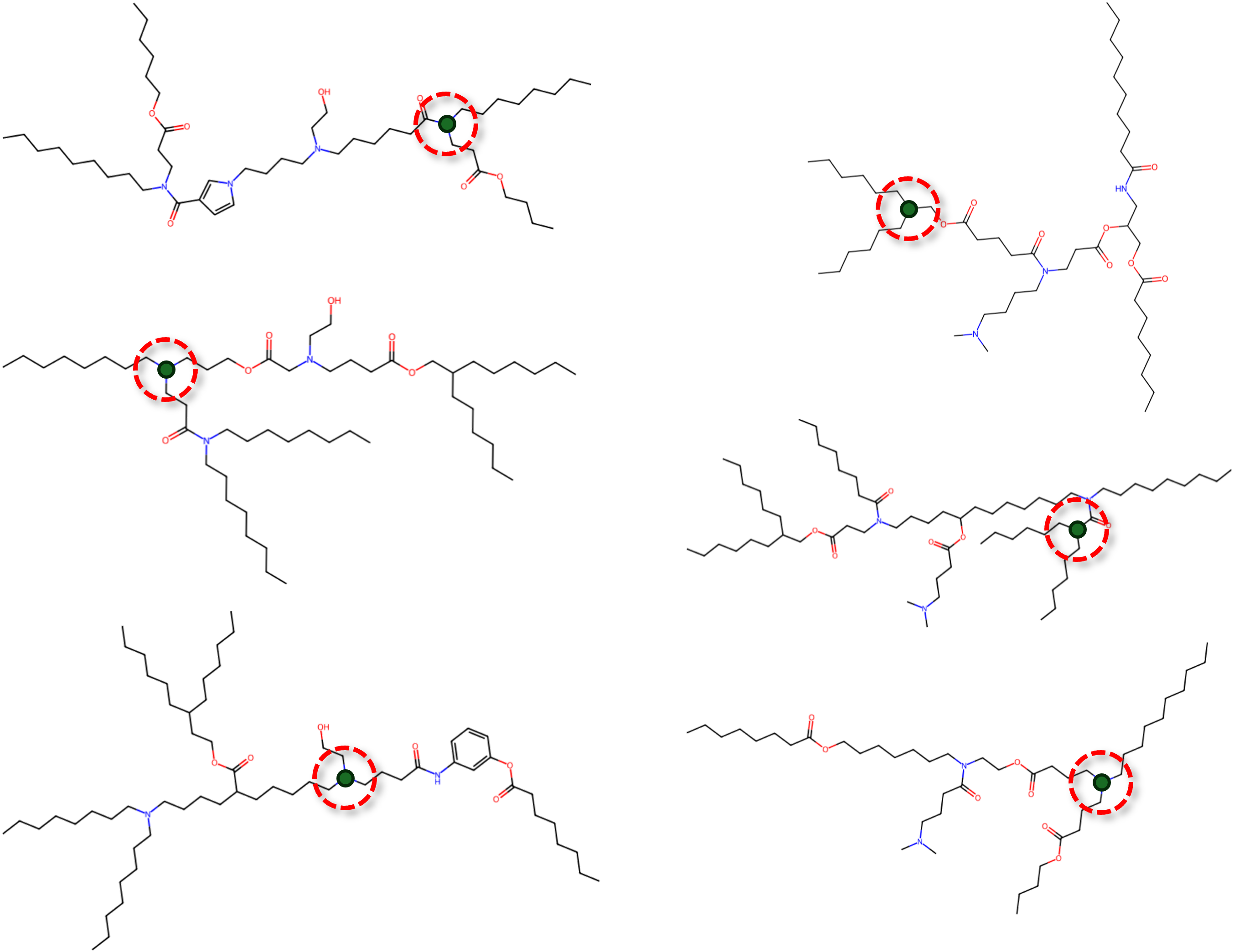}
    \caption{
        Visualization of the predicted connecting atom between head and tail via token classification, using the connecting points prediction head in the pre-trained model. The predicted atoms are explicitly marked. The visualization was created using RDKit~\cite{landrum2006rdkit}.
        }\label{fig:connecting_points_tokens}
\end{figure}

In contrast to the sequence classification method, which fails to predict all connecting atoms, the token classification model achieves a 7.4\% prediction accuracy. We selected the same six lipids as in the sequence classification model for visualizing the prediction results, as shown in Figure~\ref{fig:connecting_points_tokens}. Compared to the sequence classification method, the token classification method is able to identify "branching atoms", but it does not consistently predict the atom that connects the head and tail fragments. Among the six examples provided, only the bottom-left example correctly predicts the connecting atom. This again underscores the challenge of accurately predicting the connecting atoms.

\subsubsection{Head/Tail Classification}
The head/tail classification performs token classification on each token categorizing them into three classes: head, tail, and others. If the atom corresponding to the selected token belongs to either the head or tail, it is classified accordingly. If the token does not denote an atom (e.g., brackets, double bonds, forward/back slashes, etc.) it is classified as "others". A linear classification head is applied to transform each 768-dimensional token embeddings into a vector of size 3 for classification. The example in Figure~\ref{fig:BERT} shows the token-wise prediction labels for ALC-0315.

\begin{figure}[ht]
    \centering
    \includegraphics[width=10cm]{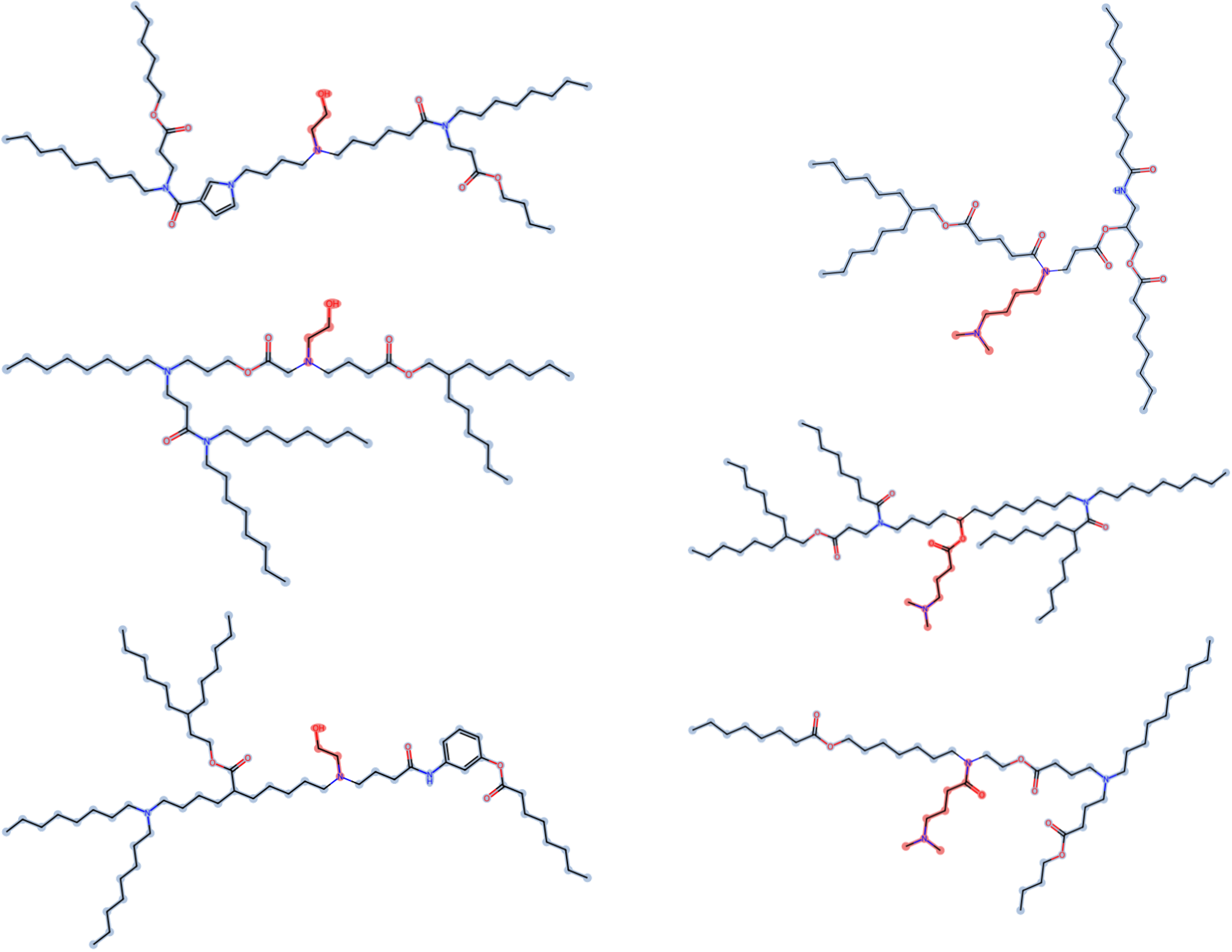}
    \caption{
        Visualization of head/tail classification using the classification head in the pre-trained model. Red represents head atoms, while blue represents tail atoms. The visualization was created using RDKit~\cite{landrum2006rdkit}.
        }\label{fig:head_tail_classification}
\end{figure}

The pre-trained classification head was used to produce the predicted head/tail labels of the randomly sampled lipids. The results are presented in Figure~\ref{fig:head_tail_classification}, where head and tail atoms are colored in red and blue, respectively, according to the predicted labels. It should be noted that thousands of lipids were checked to ensure the correct classification results, but only six lipids are shown for simplicity. For all sampled lipids, the pre-trained model achieves perfect classification of the head/tail atoms, as depicted in Figure~\ref{fig:head_tail_classification}. The automation of head/tail classification is essential in high-throughput (HTS) molecular dynamics (MD) simulations of lipid bilayer membranes. Typically, lipids consist of hydrophilic heads and hydrophobic tails. To maintain a stable bilayer membrane structure, 
the lipid heads must face the outer water layer while the tails face the inner lipid layer. Therefore, the initial positions of the lipids must be correctly placed to prevent collapse during MD simulations. The HTS lipid bilayer simulation process can be significantly accelerated by automating the head/tail labeling, compared to labor-intensive manual labeling.

\subsubsection{Rearranged/Decoy SMILES Classification}
\begin{figure}[ht]
    \centering
    \includegraphics[width=13cm]{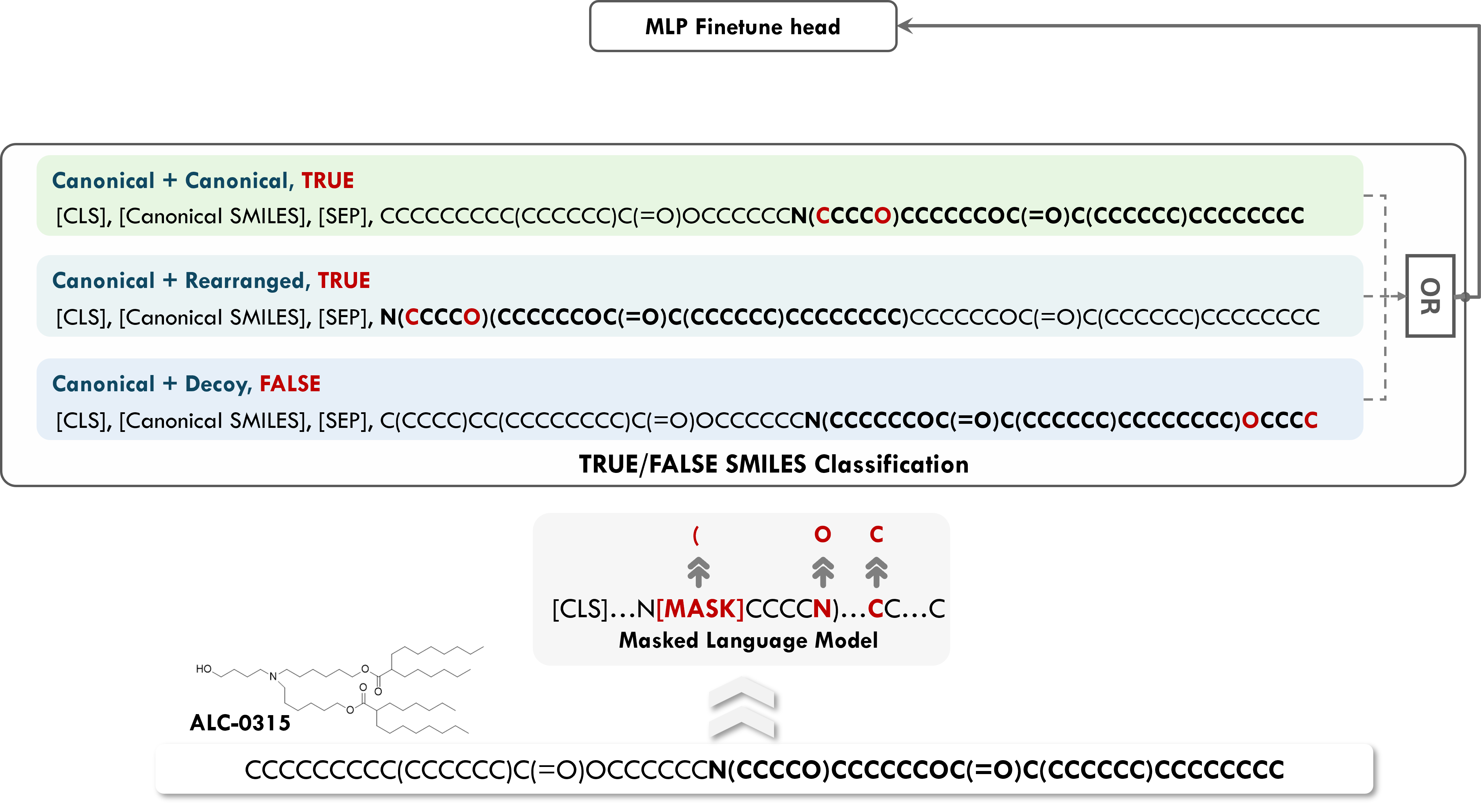}
    \caption{
        Schematic representation of the rearranged/decoy SMILES classification model.
        }\label{fig:rearranged_decoy}
\end{figure}

For a lipid molecule like ALC-0315, multiple valid SMILES starting from different atoms can be generated. However, language models like BERT~\cite{devlin2018bert} yield different embeddings for different SMILES sequences, even for distinct SMILES representing the same molecular structure. Therefore, a unified SMILES style is necessary for the LipidBERT model,
and the canonical SMILES style provided by RDKit~\cite{landrum2006rdkit} is applied in the aforementioned secondary tasks. In this secondary task, rearranged and decoy SMILES are also employed in addition to canonical SMILES, as depicted in Figure~\ref{fig:rearranged_decoy}. Rearranged SMILES refer to the SMILES that differ from the canonical SMILES but construct the same molecular structure. Such SMILES are created via RDKit by starting from a different atom. In the given example, the rearranged SMILES of ALC-0315 is created starting from the connecting nitrogen atom. Decoy SMILES are SMILES that are very similar to canonical or rearranged SMILES in their strings but actually denote structurally distinct molecules. In Figure~\ref{fig:rearranged_decoy}, the red-marked carbon and oxygen atoms are switched in the decoy SMILES to create similar but distinct molecules. The model is constructed similarly to the vanilla BERT model including MLM and NSP training objectives. The MLM model is retained exactly as in the vanilla BERT model. However, the NSP model, which predicts whether two separate sequences belong to one sentence, is replaced by the rearranged/decoy SMILES model. This new model predicts whether two SMILES refer to the same molecular structure, as depicted in Figure~\ref{fig:rearranged_decoy}. The canonical SMILES act as the first sequence and are combined with rearranged/decoy SMILES (50\% rearranged and 50\% decoy SMILES), separated by a [SEP] token. The input is labeled as "TRUE" if they represent the same molecule (rearranged) and "FALSE" if not (decoy).

In this model, we tested the classification model for several rounds, each round sampling 1,000 lipids randomly. The validation dataset was constructed in the same way as the training and test sets, with 50\% of the second sequences being rearranged SMILES and 50\% being decoy SMILES. The model performed unexpectedly well, yielding an average prediction accuracy above 99\%. These results demonstrate that the model is able to capture the structural features from lipid SMILES effectively.

\subsection{Fine-Tuning}
For each pre-trained model, we use the 768-dimensional embeddings of the [CLS] token to represent the entire sequence. The sequence output [CLS] 768-dimensional embeddings pass through a linear layer and a tanh activation function, followed by an MLP layer to generate the final output. The MLP layer consists of four layers that gradually transform the embeddings from 768 to 128 dimensions, applying the GeLU~\cite{hendrycks2016gelu} activation function. A final linear layer is employed to generate the output for regression tasks. In the context below, LipidBERT refers to the model pre-trained with or without secondary tasks that achieves the best performance (highest validation Pearson correlation coefficient value). To guard against overfitting, we applied early stopping: we recorded the maximum validation Pearson correlation coefficient (PCC) obtained over 100 training epochs. The PCC ranges from -1 to 1, where -1 represents a perfect negative linear relationship and 1 stands for a perfect positive relationship.

\subsubsection{Fine-Tuning Using Our Wet-Lab Experimental Dataset}
The up-to-date (Mar 2025) wet-lab data used for fine-tuning were provided by our experimental team and encompass a range of lipid nanoparticle (LNP) properties. Each of our pre-trained models—originally trained on datasets of varying size—was subsequently fine-tuned on these experimental LNP measurements. Below, we present representative fine-tuning results for selected LNP properties. All data in this study were generated using fixed component ratios in the LNP formulations. The primary goal of these experiments was to screen and identify the most promising ionizable lipids for downstream development of cell-, tissue-, or organ-targeted LNPs. In this manuscript, we focus on Ex-vivo organ fluorescence intensity values; for detailed results from other property models, please contact the corresponding authors.

\begin{figure}[ht]
    \centering
    \includegraphics[width=8cm]{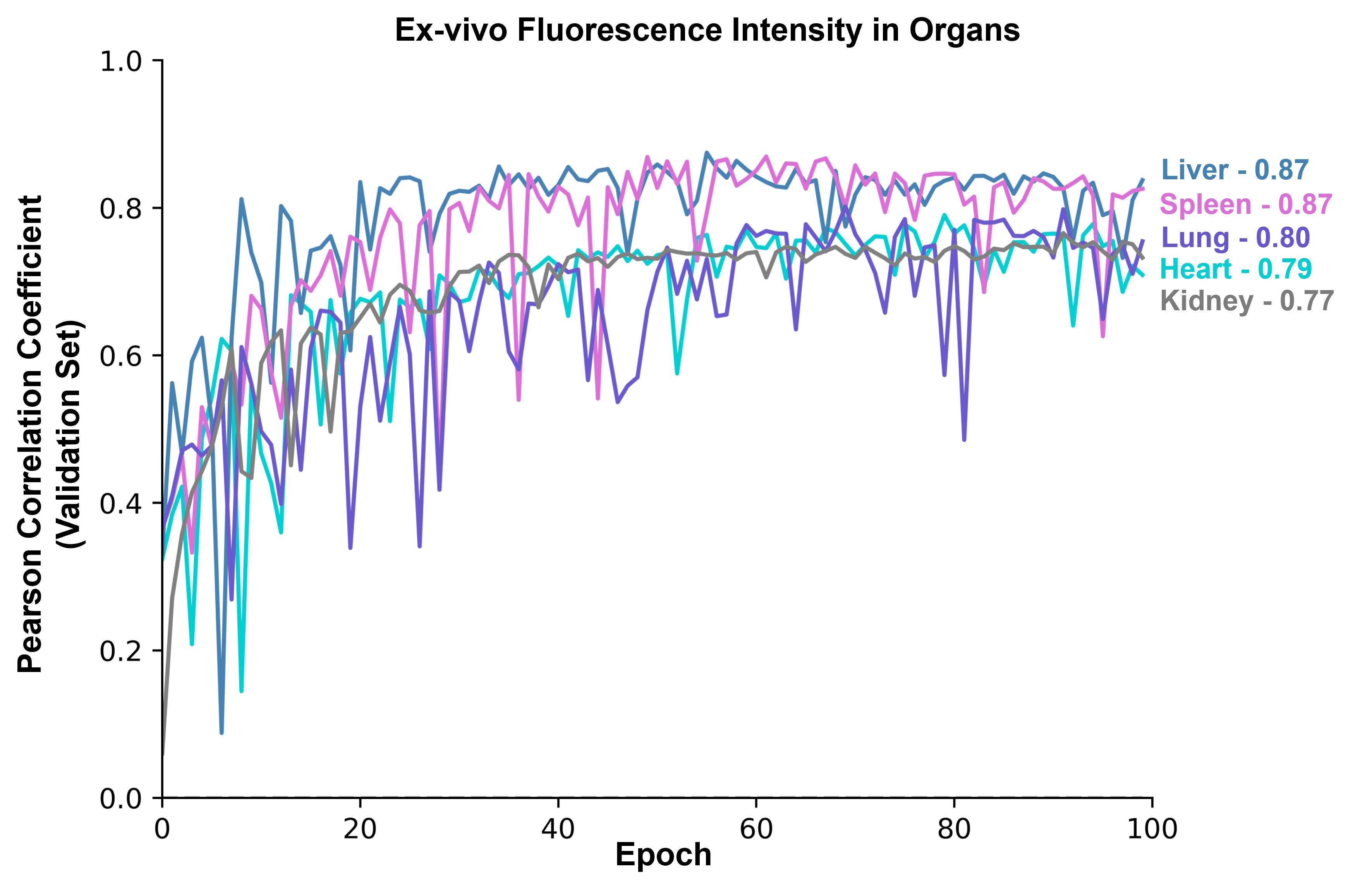}
    \caption{
        Validation set Pearson Correlation Coefficient (PCC) values change along epochs during LipidBERT fine-tuning, concerning various LNP properties.
        }\label{fig:LNP_properties_10M}
\end{figure}

\paragraph{Organ Ex-vivo Fluorescence Intensity Prediction}
First, we evaluated LipidBERT on downstream sequence-to-property regression tasks using up-to-date wet-lab fluorescence intensity (FI) measurements in five organs (heart, liver, spleen, lung, and kidney as given in Figure \ref{fig:LNP_properties_10M}). Our model achieved excellent validation Pearson correlation coefficients (PCC), all approaching or exceeding 0.80. In particular, liver and spleen converged most rapidly, stabilizing around 0.87, while the lung FI curve displayed greater variability. It is important to note that LipidBERT performs especially well on ionizable lipids whose scaffolds are well represented in our experimental set; its predictions may be less accurate for compounds with entirely novel scaffolds. As we continue to expand both the size and structural diversity of our lipid library, we expect the model’s predictive accuracy and generalizability to improve further.

\begin{figure}[ht]
    \centering
    \includegraphics[width=16cm]{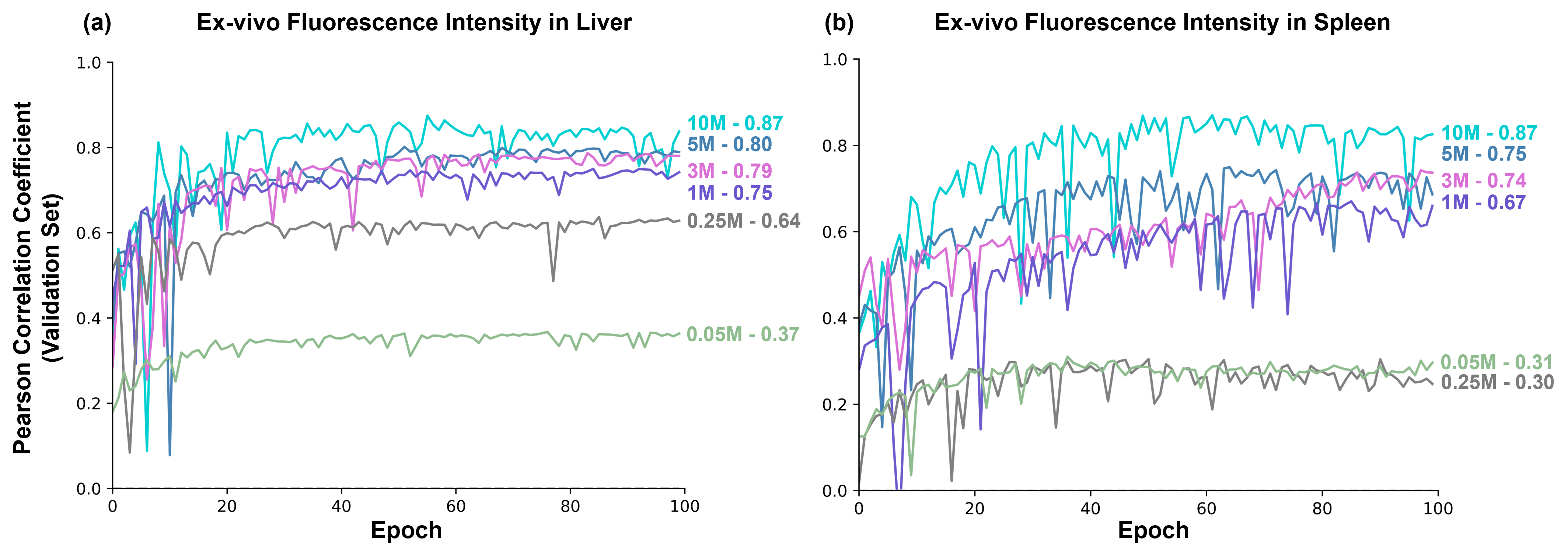}
    \caption{
        Validation set Pearson Correlation Coefficient (PCC) values change along epochs concerning (a) ex-vivo fluorescence intensity in liver, and (b) spleen. Different curves represent LipidBERT models pre-trained on different scales of training dataset.
        }\label{fig:liver_spleen_modelsize}
\end{figure}

\paragraph{Effect of Dataset Sizes}
The effect of training set sizes is also explored by fine-tuning LipidBERT models pre-trained on various training set sizes using the same experimental data mentioned above. The variation of the validation PCC values over training epochs is presented in Figure \ref{fig:liver_spleen_modelsize}. Maximum validation PCC values are selected to evaluate the performance of each model. From both figures, it is evident that model performance improves as the size of the training dataset increases. In Figure \ref{fig:liver_spleen_modelsize}(a), there is a dramatic enhancement in PCC from 0.37 with the 0.05M training set to 0.64 with the 0.25M training set, and further to 0.75 with the 1M training set. Increasing the training dataset size from 1M to 5M boosts the PCC to 0.80, while expanding the dataset from 5M to 10M yields a increase, with PCC reaching 0.87. In contrast, Figure \ref{fig:liver_spleen_modelsize}(b) shows that the 0.05M and 0.25M models exhibit negligible differences in predicting Ex-vivo spleen FI, with PCC values of 0.31 and 0.30, respectively. However, there is a significant increase in PCC between the 0.25M and 1M models, where the PCC jumps from 0.30 to 0.67. Further increases in dataset size eventually yield a 10M model with an PCC value of 0.87. These examples demonstrate that increasing the training dataset size is critical to obtaining a fine-tuned model with high PCC values, highlighting the importance of large datasets for model accuracy.

\begin{figure}[ht]
    \centering
    \includegraphics[width=13cm]{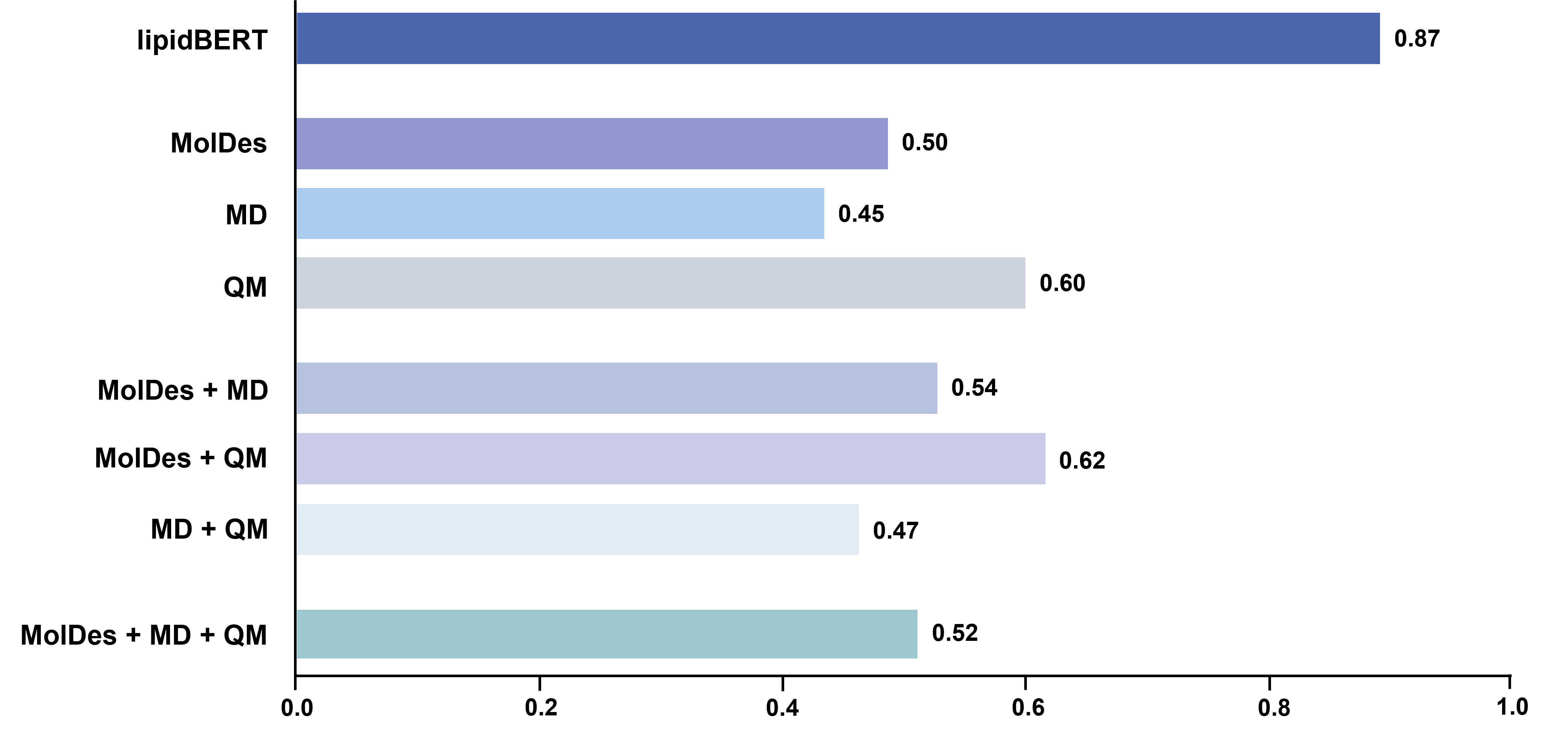}
    \caption{
        Validation set Pearson Correlation Coefficient (PCC) values of LipidBERT and various XGBoost models trained on combinations of different descriptors.
        }\label{fig:XGBoost}
\end{figure}

\paragraph{Comparison to Molecular Descriptors}
We also compared the fine-tuned 10M-LipidBERT model with our previously developed XGBoost~\cite{chen2016xgboost} model, which is trained using combinations of regular molecular descriptors, Molecular Dynamics (MD) descriptors, and Quantum Mechanics (QM) descriptors. The regular descriptors are generated via RDKit~\cite{landrum2006rdkit}, while the MD and QM descriptors are generated from MD and QM simulations, respectively. The MD descriptors are obtained from our fully-automated high-throughput (HTS) lipid bilayer simulations, designed to mimic the lipid movement in realistic LNPs. Given a lipid SMILES, the HTS platform can automatically construct the lipid/bilayer 3D structures, perform MD simulations, and extract the MD descriptors from the MD trajectories without human intervention. Similarly, the QM descriptors are also obtained automatically from our HTS platform. QM calculations are performed on a single ionizable lipid molecule to extract the QM descriptors. All descriptors are combined and used for XGBoost training on the overall fluorescence intensity, and the results are presented in Figure~\ref{fig:XGBoost}. The figure shows that the LipidBERT model exhibits overwhelming advantages over all XGBoost models, achieving an exceptionally high PCC value of 0.87. Among the XGBoost models, the QM + regular descriptors-based model gives the highest PCC value of 0.62, indicating that the combination of QM simulations and regular RDKit descriptors can derive better descriptors than other cases.

\paragraph{PhatGPT}
Apart from LipidBERT, we also trained PhatGPT (\textbf{P}recision \textbf{H}igh-throughput \textbf{A}I-driven \textbf{T}argeting using \textbf{G}enerative \textbf{P}re-trained \textbf{T}ransformer), our GPT-like \cite{radford2018GPT} lipid generation model using the same dataset as LipidBERT. Despite the fact that PhatGPT is originally designed and trained for generative tasks, we compared the performance of the two language models on downstream sequence regression tasks. It is important to note that the vanilla PhatGPT has different model parameters (such as the number of attention layers and embedding sizes). For the purpose of comparison, we pre-trained another PhatGPT model using the same parameters as LipidBERT. Several LNP properties were used for fine-tuning, but PhatGPT did not perform as expected, yielding validation PCC values around 0.3. This outcome further demonstrates the superiority of BERT-like models in sequence regression/classification tasks, while GPT-like models are primarily designed for token generation.

\paragraph{Diligent}
Diligent (\textbf{Di}ffusion‑like \textbf{Li}pid \textbf{Gen}eration \textbf{T}echnology) is our diffusion‑based generative framework trained on a dataset of 10 million lipid structures, realized in both sequence‑ and graph‑based architectures. The model leverages a DiT \cite{Peebles2022DiT} backbone, in which a transformer encoder is trained to predict the Gaussian noise at each diffusion step. In contrast to image‑based diffusion—where inputs are first partitioned into patches for pixel‑level processing—Diligent operates directly on molecular representations, diffusing either SMILES tokens or graph nodes through the forward and reverse processes. Comprehensive evaluation of these lipid generation models will be presented in a forthcoming publication.

\subsubsection{Fine-Tuning Using the AGILE Public Dataset}
The AGILE platform~\cite{xu2023agile} is an AI-Guided Ionizable Lipid Engineering platform that combines deep learning with fragment-based generative methods to accelerate the development of LNPs for mRNA delivery. The deep learning architecture of AGILE integrates graph neural networks with molecular descriptors to effectively model the complex structures and properties of ionizable lipids. Additionally, AGILE also provides publicly available transfection potency data for 1,200 lipids, tested in both Hela and Raw 264.7 cells.

\begin{figure}[ht]
    \centering
    \includegraphics[width=16cm]{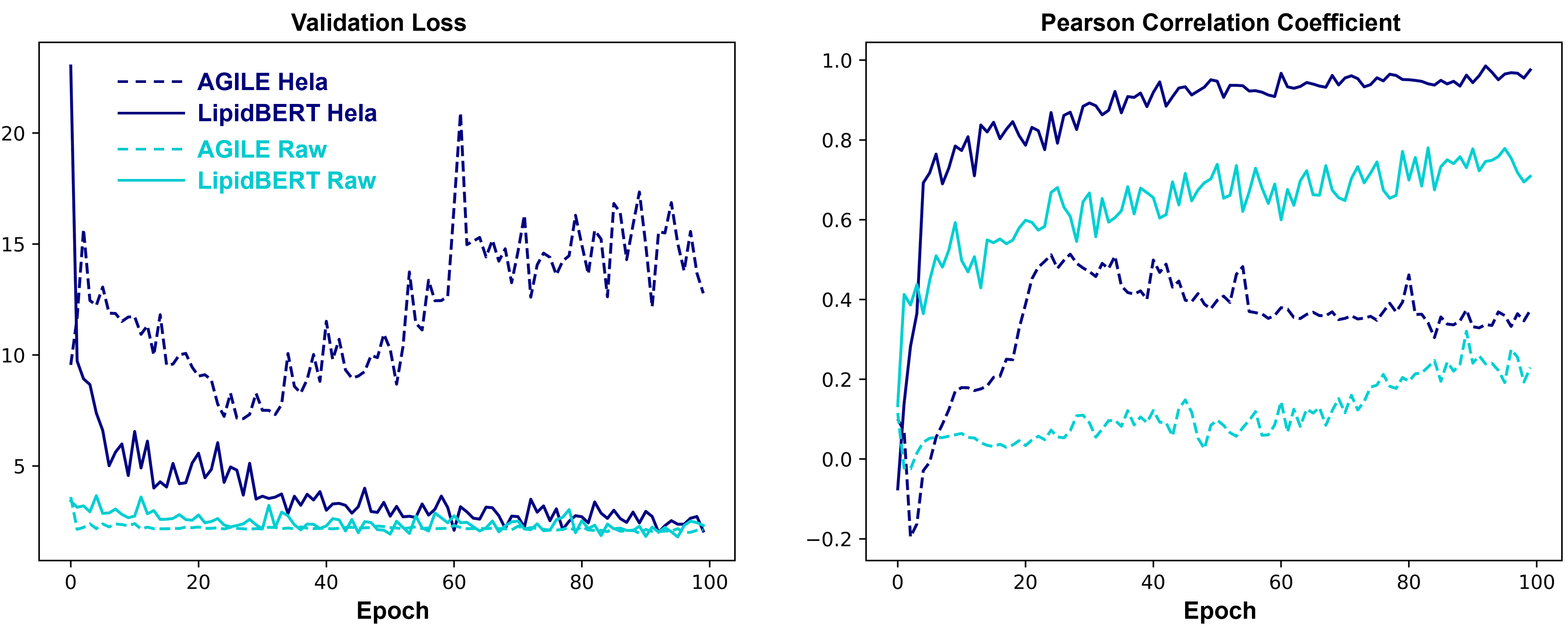}
    \caption{
        Comparison between the fine-tuned AGILE \cite{xu2023agile} and our LipidBERT model trained on the publicly available AGILE dataset.
        }\label{fig:AGILE}
\end{figure}

We followed the official instructions on AGILE's github page (\url{https://github.com/bowang-lab/AGILE}) to fine-tune the pre-trained AGILE model utilizing its provided dataset, and compared it with our fine-tuned LipidBERT model using the same dataset. Both models were fine-tuned for 100 epochs using the same random seed for an 80\%-10\%-10\% train-validation-test dataset split. The fine-tuned results, including validation loss and Pearson correlation coefficient (PCC), are given in Figure~\ref{fig:AGILE}. Generally speaking, our transformer-based LipidBERT model outperforms the graph-based AGILE model in most evaluation metrics, except for the Raw 264.7 validation loss (2.12 vs. 2.03 for AGILE and LipidBERT, respectively). For validation loss (Figure~\ref{fig:AGILE}(a)), LipidBERT experiences a rapid decline from 23 to 5.01 in the first 7 epochs, and then gradually drops to 2.07 for the Hela dataset, which is lower than the AGILE validation losses (12.8). It should be noted that the deviations between the two models concerning Raw 264.7 are negligible. In Figure~\ref{fig:AGILE}(b), all four curves start from similar Y values but end at distinctive Y values, where the LipidBERT model achieves higher values (0.98 and 0.78) than the AGILE model (0.37 and 0.23) for the Hela and Raw 264.7 datasets, respectively. Despite comparable validation losses on Raw 264.7, LipidBERT demonstrates a significantly higher correlation with ground truth. To summarize, our LipidBERT model not only exhibits state-of-the-art performance on our own experimental datasets but also on the publicly available datasets.

\subsubsection{Brief Introduction of the AiLNP Platform}
\begin{figure}[ht]
    \centering
    \includegraphics[width=16cm]{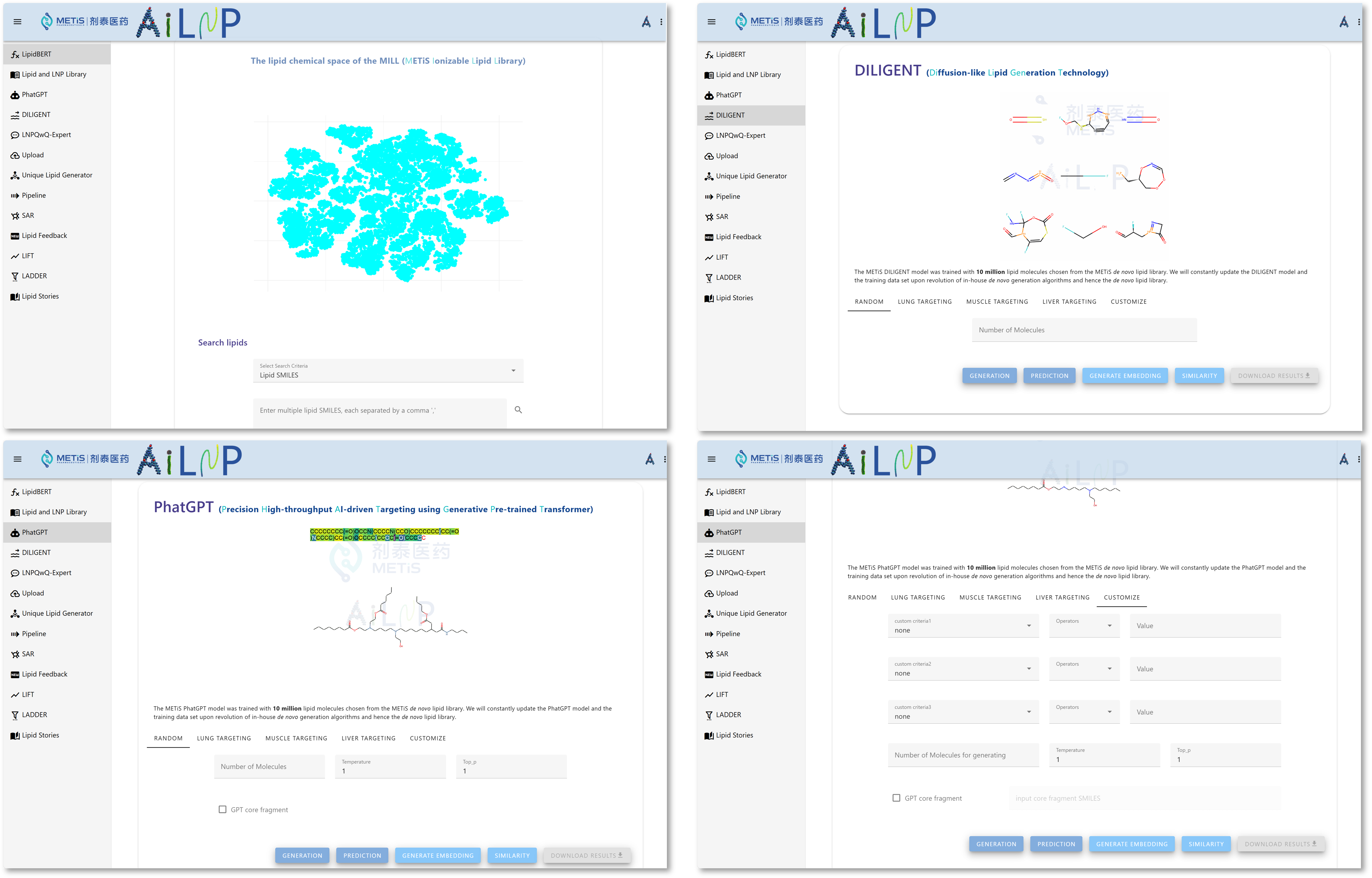}
    \caption{
        User interface of our AiLNP platform.
        }\label{fig:AiLNP}
\end{figure}

Here, we briefly present AiLNP, our AI-based platform for LNP property predictions using LipidBERT and ionizable lipid structure generation via PhatGPT. Figure \ref{fig:AiLNP} shows the front-end user interface, which incorporates a number of functions, including database searching, LNP property predictions, and lipid generation.

\section{Conclusion}
In this study, we have introduced LipidBERT, a self-supervised pre-trained language model specifically designed for \textit{in silico} generated virtual lipid molecules. Through the generation of a comprehensive database of 10 million virtual lipids using fragment-based generative methods and reinforcement learning, we have addressed the scarcity of reported lipid structures available for effective NLP pre-training. Our results demonstrate that LipidBERT, when fine-tuned for downstream lipid nanoparticle (LNP) property prediction tasks, exhibits exceptional performance with Pearson correlation coefficient values ~0.8. This underscores the efficacy of using pre-trained language models on virtual datasets for molecular property prediction.

Evidently, the METiS \textit{de novo} lipid library provides a reasonable starting point for screening and choosing potent and structurally novel lipids. This study presents a more versatile identity of the library. It can be used as a corpus for lipid representation learning, resulting in the first generation of lipid language models, which gives rise to later generations of \textit{de novo} lipid generation models. The combined use of the dry-lab work, including the virtual lipid library, molecular dynamics platform (not discussed in this study), LipidBERT and PhatGPT models, along with the wet-lab data on experimentally tested lipids and LNPs, showcases the computation-experiment collaboration and integration at the highest level. In other words, we make full use of invaluable data generated from different sources and dimensions. The high model performance, in turn, corroborates the robustness and diversity of the METiS \textit{de novo} lipid library. Therefore, the \textit{de novo} lipid library can be used more boldly and adventurously as the starting point of any AI- or DOE-assisted experimental screening of cell-, tissue-, or organ-specific lipids or LNPs, especially in the uncharted yet important territories with no existing positive ionizable lipids.

The secondary tasks designed for pre-training, such as number of tails prediction, connecting atom prediction, head/tail classification, and rearranged/decoy SMILES classification
enabled LipidBERT to gain a deep understanding of lipid structures. Despite the challenges posed by tasks like connecting atom prediction, the model demonstrated significant capability in categorizing lipids and distinguishing structural features based on SMILES sequences. Our fine-tuning experiments utilizing both proprietary and public datasets further validated LipidBERT's superior performance compared to traditional models such as XGBoost, especially in predicting LNP properties critical for mRNA delivery.

Overall, LipidBERT represents a significant advancement in the application of transformer-based architectures for the study of lipid molecules, providing a robust tool for researchers to accelerate the development of lipid-based nanomedicines. The success of LipidBERT also opens up new possibilities for the application of self-supervised learning in other areas of molecular and materials science where data scarcity has been a limiting factor.

\paragraph{Acknowledgements}
We are grateful to the METiS platform technology team for providing wet-lab experimental data. We also extend our thanks to Hongming Chen, Shaoli Liu, Yu Lu, Ruilu Feng, and Liu Yang for insightful discussions on LNP formulation. Additionally, we appreciate the valuable input from Per Larsson (Uppsala University), Michelle Lynn Hall (Eli Lilly), and Qun Zeng (XtalPi) on the applications of AI, quantum mechanics and molecular dynamics on lipid and LNP research. We thank Xin Feng for helping construct the schematic figures. We also thank Le Yin and Fei Han for useful discussions on AI and computational lipid design, as well as Wei Xu, Hongya Han, and Beibei Cao for useful discussions on biology-related topics. Our gratitude extends to Zixuan Han, Khaled Zemoura, Xiaoyun Ma, Fang Lao, Liujun Song, Kaiqi Sun, and Yang Chen for useful discussions on multiple topics. Finally, we thank Jeff Warrington for his valuable insights on the lipid library.

\newpage
\bibliographystyle{unsrt}
\bibliography{LipidBERT}

\end{document}